\documentclass[conference]{IEEEtran}
\IEEEoverridecommandlockouts
\usepackage{cite}
\usepackage{amsmath,amssymb,amsfonts}
\usepackage{algorithmic}
\usepackage{graphicx}
\usepackage{textcomp}
\usepackage{xcolor}

\usepackage{multirow}
\usepackage{booktabs} 
\usepackage{makecell}
\usepackage{enumitem}

\usepackage{amsmath}
\usepackage{amsthm}
\usepackage{amsfonts}
\usepackage{hyperref}
\usepackage{placeins}
\usepackage{float}

\usepackage{authblk}

\def\BibTeX{{\rm B\kern-.05em{\sc i\kern-.025em b}\kern-.08em
    T\kern-.1667em\lower.7ex\hbox{E}\kern-.125emX}}
\begin{document}

\title{ZeroKBC: A Comprehensive Benchmark for \\ Zero-Shot Knowledge Base Completion}

\author[1]{Pei Chen}
\author[2]{Wenlin Yao}
\author[2]{Hongming Zhang}
\author[2]{Xiaoman Pan}
\author[2]{Dian Yu}
\author[2]{Dong Yu}
\author[2]{Jianshu Chen}
\affil[1]{Department of Computer Science and Engineering, Texas A\&M University
}
\affil[1]{\texttt{chenpei@tamu.edu}}
\affil[2]{Tencent AI Lab
}
\affil[2]{\texttt{\{wenlinyao,hongmzhang,xiaomanpan,yudian,dyu,jianshuchen\}@global.tencent.com}}
\renewcommand\Authands{ and }

\maketitle

\begin{abstract}
Knowledge base completion (KBC) aims to predict the missing links in knowledge graphs. Previous KBC tasks and approaches mainly focus on the setting where all test entities and relations have appeared in the training set. However, there has been limited research on the zero-shot KBC settings, where we need to deal with unseen entities and relations that emerge in a constantly growing knowledge base. In this work, we systematically examine different possible scenarios of zero-shot KBC and develop a comprehensive benchmark, ZeroKBC, that covers these scenarios with diverse types of knowledge sources. Our systematic analysis reveals several missing yet important zero-shot KBC settings. Experimental results show that canonical and state-of-the-art KBC systems cannot achieve satisfactory performance on this challenging benchmark. By analyzing the strength and weaknesses of these systems on solving ZeroKBC, we further present several important observations and promising future directions.\footnote{Work was done during the internship at Tencent AI lab. The data and code are available at: \url{https://github.com/brickee/ZeroKBC}}     
\end{abstract}

\begin{IEEEkeywords}
zero-shot, knowledge base completion, benchmark
\end{IEEEkeywords}

\section{Introduction}


Knowledge bases are important resources for applications such as question answering, commonsense reasoning, and machine reading comprehension. They are usually represented as knowledge graphs (KGs), which organize knowledge in a structured form with nodes being entities and edges being the relations between entities. Generally, there are three types of KGs~\cite{Storks2019RecentAI}: \textit{linguistic knowledge} such as WordNet~\cite{miller1995wordnet}, \textit{world knowledge} such as Freebase~\cite{bollacker2008freebase}, and \textit{commonsense knowledge} such as ConceptNet~\cite{speer2017conceptnet}. One of the biggest bottlenecks of these KGs is that they are highly incomplete. So the task of \textit{knowledge base completion} (KBC) (a.k.a \textit{link prediction}) that aims to infer the missing links in the graphs has gained much research attention~\cite{taskar2003link}.

However, most previous KBC tasks assume that test entities and relations have been seen during training. In practical scenarios, a KG is constantly growing over time, and new entities and relations will be continuously added to an existing KG. As a result, a practical KBC system will have to deal with such a zero-shot KBC setting, where certain entities and relations have not been seen during the training time. Recent work has studied zero-shot KBC scenarios with either unseen entities~\cite{sun-etal-2020-evaluation,baek2020gen} or unseen relations~\cite{qin2020generative} individually (but not both). However, there is still no comprehensive zero-shot KBC benchmark that systematically studies different scenarios in the zero-shot KBC settings with different combinations of available information.

\label{sec:appendix_b}
\begin{table*}[h!]
\centering
\small
\setlength{\tabcolsep}{4.pt}
\caption{Scenario Coverage Comparison (Previous Work v.s Our Work). }
\begin{tabular}{|c|c|c| c| c| c |c |c|c| c| c |c |c|c| c|c |}

\hline
\multirow{2}{*}{Related Work} & \multicolumn{12}{|c|}{Settings} & \multicolumn{3}{|c|}{Knowledge Sources} 
\\

\cline{2-16}
  &  1 & 2 & 3 & 4 & 5 & 6  & 7  & 8 & 9 & 10  & 11  & 12  & Linguistic & World& Commonsense \\
 \hline
TransE \cite{bordes2013translating} &  &  &  & \checkmark   &  &  &   &    &  &  &  &  & \checkmark&  \checkmark& \\
 \hline

 \cite{neelakantan-etal-2015-compositional} &  &  &  &   &  &  &   &     &  & \checkmark &  &  & &  \checkmark& \\
 \hline
 
  \cite{xie2016representation} &  & \checkmark &  &   &  & \checkmark &   &     &  &  &  &  & &  \checkmark& \\
 \hline
 
 \cite{ijcai2017-0250} &  &  &  &   &  &  &   &   \checkmark  &  &  &  &  & \checkmark&  \checkmark& \\
 \hline
 
 RotatE \cite{sun2018rotate} &  &  &  & \checkmark   &  &  &   &    &  &  &  &  & \checkmark& \checkmark& \\
 \hline

\cite{albooyeh-etal-2020-sample} &  &  &  &   &  &  &   &   \checkmark  &  &  &  &  & \checkmark&  \checkmark& \\
 \hline
 
 \cite{baek2020gen} &  &  &  &   &  &  &   &   \checkmark  &  &  &  &  & \checkmark&  \checkmark& \\
 \hline
 
 \cite{malaviya2020commonsense} &  &  \checkmark &  &   &  &  &   &     &  &  &  &  & &  & \checkmark\\
 \hline
\cite{qin2020generative} &  &  &  &    &  &  &   &    &  & \checkmark  &  &  & \checkmark& \checkmark& \\
 \hline
 
 StAR \cite{wang2021structure} & \checkmark &  &  &    &  &  &   &    &  &  &  &  & \checkmark& \checkmark& \\

 \hline
Our Work & \checkmark & \checkmark  & \checkmark &  \checkmark  &\checkmark  &\checkmark  & \checkmark  &  \checkmark  &\checkmark  & \checkmark &\checkmark  & \checkmark & \checkmark& \checkmark& \checkmark \\

\hline

\end{tabular}
\label{tab:full_settings}
\end{table*}

\begin{figure*}[t]
\setlength{\belowcaptionskip}{-15pt}
	\begin{center}
	\includegraphics[width=500pt]{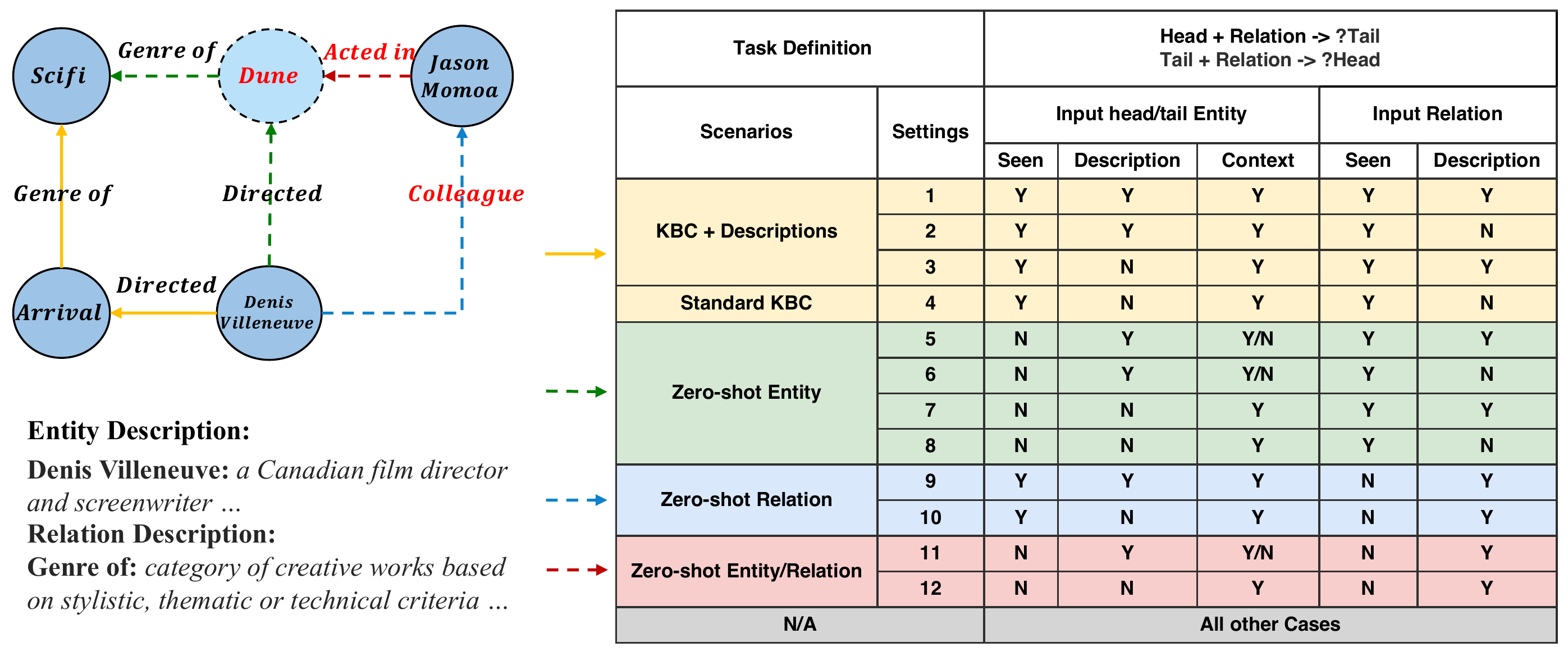}
	\caption{\textbf{Knowledge Graph Sample (Left):} Black names of entities and relations  represent "seen" and red ones mean "unseen" or "newly emerged". Differently colored arrows in the graph correspond to different scenarios on the right. \textbf{KBC task on different scenarios (Right):} "Y" means "yes" or "required", "N" means "no" or "not required" and "Y/N" means "optional". The colored arrows correspond to the link prediction scenarios in the left graph sample.
	\textbf{Context} means known connections for an unseen entity as explained in Section~\ref{section:zero_kbc}.
}
	\label{fig:task}
	\end{center}
\end{figure*}

For this reason, we propose a new benchmark, \textbf{ZeroKBC}, which covers all the useful zero-shot KBC scenarios with three different types of knowledge resources. In total, we have $3$ zero-shot scenarios based on whether the input entity and relation are unseen, and they are further categorized into $8$ fine-grained settings based on the availability of the textual descriptions or context in the input (Figure~\ref{fig:task}). 
To our best knowledge, our work is the first to systematically categorize all possible zero-shot KBC settings. And because of it, we are able to identify $5$ practically important zero-shot KBC settings that were missing in previous studies (Table \ref{tab:full_settings}).
Specifically, each previous work in Table \ref{tab:full_settings} only covers a small portion of all KBC settings, while our work has more exhaustive coverage. 
Furthermore, because of our systematic categorization and analysis, we found that there are $6$ important KBC settings that were missing in previous study, i.e., standard setting \#3 and zero-shot settings \#5, \#7, \#9, \#11 and \#12. These are the scenarios of important value, but could be easily ignored without such a systematic analysis of all the possible KBC settings.
We adapt state-of-the-art KBC systems to these zero-shot scenarios, and the results show that ZeroKBC is quite challenging and far from being solved. Our analysis further reveals the importance of designing more efficient methods to integrate semantics of entities and relations into knowledge graph learning.

\section{Related Work}

KBC benchmarks mostly assume that all test entities and relations appear during training, and many structure-based methods are proposed for this scenario~\cite{bordes2013translating,dettmers2018conve,Nguyen2018,sun2018rotate,nathani-etal-2019-learning,8946600}. Despite of just using the graph structure, some recent work~\cite{xie2016representation,malaviya2020commonsense,wang2021structure} found that incorporating the text descriptions of entities and relations can further improve the performance.



Recently, emerging entities~\cite{ijcai2017-0250,albooyeh-etal-2020-sample,baek2020gen} and relations~\cite{neelakantan-etal-2015-compositional,chen-etal-2019-meta,qin2020generative} have also caught research attentions. However, comprehensive KBC settings are seldom explored, and performance are incomparable because of different benchmark and evaluation protocols. This work systematically analyzes all possible KB growing scenarios and constructs ZeroKBC. We identify $5$ more zero-shot settings that merit important application values (Table \ref{tab:full_settings}).





\section{Knowledge Base Completion Task}
\label{section:kbc}
We first formally define the knowledge graph (KG) and the standard KBC task.  
\paragraph{Knowledge Graph}
Let $\mathcal{E}$ be the entity set and $\mathcal{R}$ be the relation set. A \textbf{triple} is defined as $(e_h, r, e_t)$ where $e_h, e_t\in \mathcal{E}$ are the head and the tail entities, respectively. A knowledge graph is denoted as $\mathcal{G} = \{(e_h, r, e_t)\} \subseteq \mathcal{E}\times \mathcal{R}\times \mathcal{E} $. Meanwhile, 
each entity $e$ may have a text description $d_e=\left\{w_{1:n}\right\}$ with $n$ words, and each relation $r$ may have a text description $d_r=\left\{w_{1:m}\right\}$ with $m$ words.
\paragraph{KBC}
It aims to predict the \textbf{missing} triples $(e_h, r, ?)$ or $(?, r, e_t)$ given $\mathcal{G}$ when all the entities $e_h, e_t, ? \in \mathcal{E}$ and relation $r \in \mathcal{R}$ have been seen in other triples during training. 
The standard KBC setting only considers structural information of the knowledge graph to make predictions (setting 4 in Figure \ref{fig:task}).
For example, as shown in Figure~\ref{fig:task}, predicting the "Genre of" relation for entities "Sci-fi" and "Arrival" is a standard KBC task as long as "Genre of" also occurs in the training set. 
Also, textual descriptions of entities and relations 
can be utilized 
for performance enhancement (settings 1--3 in Figure \ref{fig:task}). 

\section{Zero-Shot Knowledge Base Completion}
\label{section:zero_kbc}

In Figure~\ref{fig:task}, we systematically categorize 3 zero-shot scenarios with 8 fine-grained settings. Because of our systematic categorization and analysis, we found that there are $5$ important zero-shot KBC settings that were missing in previous studies, i.e., settings \#5, \#7, \#9, \#11 and \#12. These are the scenarios of important value, but could be easily ignored without such an exhaustic analysis (see Table \ref{tab:full_settings} for a comprehensive comparision with existing works). Next, we discuss the 3 macro-categories of zero-shot scenarios (represented by the green, blue and red color in Figure~\ref{fig:task}).
Next we define zero-shot KBC scenarios when new entities or relations emerge during testing. 

\paragraph{Zero-Shot Entity KBC (ZeroE)}
Given a graph $\mathcal{G} \subseteq \mathcal{E}\times \mathcal{R}\times \mathcal{E}$, a newly emerged entity is an entity $\tilde{e} \in \tilde{\mathcal{E}}$ where $\mathcal{E} \cap \tilde{\mathcal{E}}=\emptyset$. When such a new entity is added to the existing KG, we need to predict all the possible links between this new entity and existing entities, i.e., $(\tilde{e}_h, r, ?)$ or $(?, r, \tilde{e}_t)$, where $\tilde{e}_h, \tilde{e}_t \in \tilde{\mathcal{E}}, ? \in \mathcal{E}$, and relation $r \in \mathcal{R}$. For the new entity $\tilde{e}$, it is possible that we already know some \textbf{context}\footnotemark[2] for it, i.e., we know some connections $(\tilde{e}, r, e_t)$ or $(e_h, r, \tilde{e})$ between this new entity and the existing KG. We can utilize such context to help predict the remaining missing links. In the absence of such context for a new entity, we need the textual descriptions of it to distinguish different entities. For example, as in Figure~\ref{fig:task}, predicting the ``Directed'' relation for the new unseen entity ``Dune'' and existing entity ``Denis Villeneuve'' is a ZeroE task. We can either utilize the context (Dune, Genre of, Sci-fi) or the textual description of the new entity ``Dune'' for prediction.

\footnotetext[2]{This does not mean that $\tilde{e}$ has been seen because such context is not observed during training, as pointed out by \cite{albooyeh-etal-2020-sample}.}

\paragraph{Zero-Shot Relation KBC (ZeroR)}
Given a graph $\mathcal{G} \subseteq \mathcal{E}\times \mathcal{R}\times \mathcal{E}$, a newly emerged relation is a relation $\tilde{r} \in \tilde{\mathcal{R}}$ where $\mathcal{R} \cap \tilde{\mathcal{R}}=\emptyset$. When such a relation is added to the existing KG, we need to predict if this relation exists in any pair of the entities, i.e., $(e_h, \tilde{r}, ?)$ or $(?, \tilde{r}, e_t)$ where $e_h, e_t, ? \in \mathcal{E}$, and relation $\tilde{r} \in \tilde{\mathcal{R}}$. In this scenario, we will always have the textual descriptions of new relations to distinguish them since we have never seen them before. For instance, predicting the ``Colleague'' relation for the entities ``Denis Villeneuve'' and ``Jason Momoa'' from Figure~\ref{fig:task} belongs to this setting.

\paragraph{Zero-Shot Both KBC (ZeroB)}
It is also possible that we encounter the scenario to predict the link for both a newly added relation and a newly added entity, i.e., $(\tilde{e}_h, \tilde{r}, ?)$ or $(?, \tilde{r}, \tilde{e}_t)$ where $\tilde{e}_h, \tilde{e}_t \in \tilde{\mathcal{E}}, ? \in \mathcal{E}$, and relation $\tilde{r} \in \tilde{\mathcal{R}}$. 
Meanwhile, $\mathcal{E} \cap \tilde{\mathcal{E}}=\emptyset$, and $\mathcal{R} \cap \tilde{\mathcal{R}}=\emptyset$.
In such a case, textual description is necessary for the relations as in ZeroR and entity descriptions are also required in the absence of context (Setting 12). Predicting the triple (Jason Momoa, Acted in, Dune) in Figure~\ref{fig:task} is a Zero-shot Both KBC case.

\section{A Comprehensive Benchmark Dataset}

\begin{table}[t]
\setlength{\tabcolsep}{1.1pt}
\centering
\caption{Statistics of ZeroKBC (Un.: unseen, Avg.: average, Wrd.: word, Ent: entity, and Rel.: relation).}
	\begin{tabular}{@{}l|l|r|r|r}
		\toprule
		\begin{tabular}[c]{@{}l@{}} Scenarios \end{tabular} &  \multicolumn{1}{c|}{Statistics}  & \begin{tabular}[c]{@{}l@{}}   \makecell{WN18RR} \end{tabular} &
		\begin{tabular}[c]{@{}l@{}}  \makecell{FB15K237} \end{tabular} &
		\begin{tabular}[c]{@{}l@{}}  \makecell{ATOMIC}\end{tabular}

		\\ \midrule
		\multirow{5}{*}{Standard}     & \# All Triples &    93,003 & 310,116 & 389,437 \\  
	   & \# Ent. & 40,943 &14,541 & 48,645	  \\ 
	   & \# Rel.  & 11 & 237 & 9    \       \\   
	   & \# Avg. Wrd. Ent. & 14.24 & 7.34 & 3.72\       \\
	   & \# Avg. Wrd. Rel. & 2.55 & 19.24 & 4.44\       \\
	   \midrule
	   \multirow{3}{*}{ZeroE}     & \# Seen Ent.  &    32,270 & 11,579 &  38,903  \\  
	   & \# Un. Dev Ent. & 2,848  &  1,395&  4,768 \\ 
	   & \# Un. Test Ent.  & 2,848 &  1,396& 4,691\       \\   \midrule
	   \multirow{3}{*}{ZeroR}     & \# Seen Rel. &    6 & 157 &  5  \\  
	   & \# Un. Dev Rel. &    2 & 20 &  2 \\ 
	   & \# Un. Test Rel.  &    3 & 60 & 2\       \\   \midrule
	   \multirow{6}{*}{ZeroB}     & \# Seen Ent.  &   30,015  & 11,252 &   32,919 \\  
	   & \# Un. Dev Ent. & 54 & 449&  2,577\\ 
	   & \# Un. Test Ent.  & 225 & 962& 2,977\       \\   
	   & \# Seen Rel. & 6 & 154 &  5\\  
	   & \# Un. Dev Rel. & 2 & 19 & 2  \\ 
	   & \# Un. Test Rel.  & 3 & 59 &2       \\   

		\bottomrule
	\end{tabular}
\label{tab:stats}
\end{table}

\begin{table*}[htp]
\centering
\setlength{\tabcolsep}{3pt}
\caption{Main results on ZeroKBC (\% for MRR and Hits@X). Full results for all 12 settings are given in Table \ref{tab:full_results}. By using the unified evaluation schema, the results of standard KBC (setting 4) are different from previous work.
}
\begin{tabular}{|c|c|c| r| r| r |r |r| r| r| r| r| r|r| r| r| r| r|}

\hline
\multirow[b]{2}{*}{Scenarios} & \multirow[b]{2}{*}{Settings} &\multirow[b]{2}{*}{Models} & \multicolumn{5}{|c|}{WN18RR} & \multicolumn{5}{|c|}{FB15K237} & \multicolumn{5}{|c|}{ATOMIC} \\
\cline{4-18}
  &  & & \multicolumn{1}{|c|}{MR} & \multicolumn{1}{|c|}{MRR} & \makecell{Hits\\ @10} & \makecell{Hits\\ @3}  & \makecell{Hits\\ @1}  & \multicolumn{1}{|c|}{MR} & \multicolumn{1}{|c|}{MRR} & \makecell{Hits\\ @10}  & \makecell{Hits\\ @3}  & \makecell{Hits\\ @1}  & \multicolumn{1}{|c|}{MR} & \multicolumn{1}{|c|}{MRR} & \makecell{Hits\\ @10} & \makecell{Hits\\ @3}  & \makecell{Hits\\ @1} 

\tabularnewline
\hline
\multirow{3}{*} {Standard}   & \multirow{3}{*} {1}   & TransE & 2368
 & 20.8 & 50.8 & 36.4 & 1.7 & 193  & 31.5  & 50.3 & 35.0 & 22.0 & 2442 & 20.2 & 29.6 & 22.2 & 15.2  \tabularnewline
 & & RotatE  & 2814 & \textbf{42.6} & 54.0 & \textbf{46.1} & \textbf{36.1} & 193 & \textbf{32.2} & \textbf{51.5}  & \textbf{35.8} & \textbf{22.7} & 2318 & \textbf{24.1} & \textbf{32.6} & \textbf{25.2}& \textbf{19.8} \tabularnewline
 
 &  & StAR &  \textbf{68} & 34.6 & \textbf{62.3} & 41.9 & 20.3 &  \textbf{127} & 27.4 & 45.8 & 30.0 & 18.4
 & \textbf{1566}  & 7.8 & 12.5 & 7.5 & 5.2
 
 \tabularnewline
 \hline
\multirow{3}{*} {ZeroE} & \multirow{3}{*} {\makecell{5\\ w/ context}} & TransE & 3598  & 7.6 & 12.1  & 7.9 & 5.1  & 787 & 11.8 & 19.9  & 12.4 & 7.7 & 6170 & 0.8 & 1.5 & 0.7 & 0.3 \tabularnewline
                          & & RotatE  &  4064 & 7.6 & 11.6 & 7.8 & 5.4 & 2412 & 2.6  & 4.9 & 2.5 & 1.1 & 11872 & 0.2 & 0.4 & 0.2 & 0.1 \tabularnewline
                          & & StAR   &  \textbf{487} & \textbf{36.3} & \textbf{58.0} & \textbf{41.0} & \textbf{25.5} & \textbf{236} & \textbf{22.5} & \textbf{38.7}  & \textbf{24.2} & \textbf{14.5} &  \textbf{1399} & \textbf{5.7} & \textbf{10.5} & \textbf{5.3} & \textbf{3.0}
 \tabularnewline
 \hline
\multirow{3}{*} {ZeroR} & \multirow{3}{*} {9}  & TransE & 19584 &   0.3 & 0.4 & 0.2 &  0.1 & 6922 & 2.7 & 4.3 & 2.8 & 1.7 & 7799 & 0.5  & 0.8 & 0.2 & 0.1 \tabularnewline
                         & & RotatE  & 21876 & 0.2 & 0.4 & 0.2 & 0.1 & 6645 & 1.6 & 2.3 & 1.6 & 1.2 & 18562 & 0.1 & 0.1 & 0.0 & 0.0 \tabularnewline
                         & & StAR   & \textbf{2810} & \textbf{9.7} & \textbf{24.6} & \textbf{11.1} & \textbf{2.4}  & \textbf{1836} & \textbf{11.8} & \textbf{21.2} & \textbf{12.1} & \textbf{7.0} &  \textbf{2787} & \textbf{2.6} & \textbf{4.3} & \textbf{2.3} & \textbf{1.4}
\tabularnewline
 \hline
\multirow{3}{*} {ZeroB}   & \multirow{3}{*} {\makecell{11\\ w/ context}} & TransE &  \textbf{4830} & 4.0 & 9.0 & 3.2 & 2.1 & 3581 & 3.8 & 6.9 & 4.2 & 1.9 & 15032 & 2.2 & 3.8 & \textbf{2.0} & 0.1 \tabularnewline
                         & & RotatE  & 7116 & 4.3 & 8.9 & 4.1 & 2.6 & 3855 & 1.5 & 3.8 & 1.1 & 0.4  & 10405 & 1.8 & 1.9 & 1.6 & \textbf{1.6} \tabularnewline
                         & & StAR  & 5317 & \textbf{14.1} & \textbf{29.6} & \textbf{15.4} & \textbf{7.7} & \textbf{1561} & \textbf{10.4} & \textbf{19.2} & \textbf{10.5} & \textbf{5.8} &  \textbf{2102} & \textbf{2.3} & \textbf{4.1} & \textbf{2.0} & 1.0
 \tabularnewline

 \hline
\end{tabular}

\label{tab:main_results}
\end{table*}




\paragraph{Dataset Preparation} Considering all the KBC scenarios, we propose a comprehensive benchmark ZeroKBC that includes three types of knowledge resources: linguistic, world, and commonsense knowledge~\cite{Storks2019RecentAI}. For linguistic knowledge, we consider WN18RR~\cite{dettmers2018convolutional}, a subset of WordNet that includes eleven relations; for world knowledge, we include FB15k-237~\cite{toutanova-chen-2015-observed}, a subset of Freebase that includes 15 thousand entities and 237 relations; we use ATOMIC~\cite{sap2019atomic} that includes nine relations for commonsense knowledge. All three resources have textual descriptions for both entities and relations. 
Specifically, we consider three zero-shot scenarios: ZeroE, ZeroR, and ZeroB as follows.
For the ZeroE scenario, we follow the dataset construction procedures introduced by~\cite{albooyeh-etal-2020-sample} and use the out-of-sample entities as the unseen entities for evaluation; for the ZeroR scenario, we follow~\cite{qin2020generative} for construction and set aside a part of relations for evaluation only; as for the ZeroB scenario, we take the intersection of the previous two zero-shot datasets. The statistics of the whole zero-shot benchmark dataset are shown in Table~\ref{tab:stats}.

\paragraph{Unified Evaluation Schema} 
As pointed out by~\cite{sun-etal-2020-evaluation}, many previous KBC systems use an inappropriate evaluation protocol and report untrustworthy performance. When the predicted scores are the same for the candidates, they rank the correct one on the top and substantially overestimate the final
ranking scores. Thus, we unify the evaluation schema in all our experiments and rank the correct one in the middle in such cases. Besides, for the ZeroE and ZeroB scenarios, we adopt the protocol from~\cite{albooyeh-etal-2020-sample} and only consider the seen entities as candidates for ranking. We also adopt the standard filtering strategy introduced by~\cite{bordes2013translating}.  







\begin{table*}[h!]
\centering
\setlength{\tabcolsep}{2.2pt}
\small
\caption{Complete Experimental Results on ZeroKBC (\% for MRR and Hits@X).}
\begin{tabular}{|c|c|c| r| r| r |r |r|r| r| r |r |r|r| r| r |r |r|}

\hline
\multirow[b]{2}{*}{Scenarios}&\multirow[b]{2}{*}{Settings} & \multirow[b]{2}{*}{Models} & \multicolumn{5}{|c|}{WN18RR} & \multicolumn{5}{|c|}{FB15K237} & \multicolumn{5}{|c|}{ATOMIC} \\
\cline{4-18}
  &  & & \multicolumn{1}{|c|}{MR} & \multicolumn{1}{|c|}{MRR} & \makecell{Hits\\ @10} & \makecell{Hits\\ @3}  & \makecell{Hits\\ @1}  & \multicolumn{1}{|c|}{MR} & \multicolumn{1}{|c|}{MRR} & \makecell{Hits\\ @10}  & \makecell{Hits\\ @3}  & \makecell{Hits\\ @1}  & \multicolumn{1}{|c|}{MR} & \multicolumn{1}{|c|}{MRR} & \makecell{Hits\\ @10} & \makecell{Hits\\ @3}  & \makecell{Hits\\ @1} 

\tabularnewline
\hline
\multirow{12}{*} {Standard}  & \multirow{3}{*} {1} & TransE  & 2368 & 20.8 & 50.8 & 36.4 & 1.7 &  193 & 31.5 & 50.3 & 35.0 & 22.0
  & 2442 & 20.2 & 29.6 & 22.2 & 15.2 \tabularnewline
&  & RotatE &  2814 & 42.6 & 54.0 & 46.1 & 36.1 &  193 & 32.2 & 51.5 & 35.8 & 22.7 & 2318 & 24.1 & 32.6 & 25.2 & 19.8 \tabularnewline
&  & StAR &  68 & 34.6 & 62.3 & 41.9 & 20.3 &  127 & 27.4 & 45.8 & 30.0 & 18.4
 &  1566 & 7.8 & 12.5 & 7.5 & 5.2
 \tabularnewline
\cline{2-18}

& \multirow{3}{*} {2} & TransE  & 2309 & 20.8 & 50.7 & 37.2 & 1.0 &  191 & 31.2 & 50.5 & 35.0 & 21.5 & 2390 & 21.2 & 30.3 & 22.7 &16.5
\tabularnewline
&  & RotatE &  2799 & 42.9 & 53.8 & 46.4 & 36.6
 &  192 & 32.2 & 51.5 & 36.0 & 22.7
 &  2328 & 24.2 & 32.5 & 25.5 & 19.9
 \tabularnewline
&  & StAR &  34 & 74.7 & 62.4 & 40.7 & 18.9
 &  128 & 27.6 & 46.4 & 30.2 & 18.4
 &  1566 & 8.1 & 12.6 & 7.9 & 5.6
\tabularnewline
\cline{2-18}

& \multirow{3}{*} {3} & TransE  & 3800 & 20.0 &  48.8 & 33.5 & 2.8 &  189 & 31.9 & 51.1 & 35.6 & 22.3 &  2516 & 24.4 & 32.5 & 25.7 & 20.1
\tabularnewline
&  & RotatE &  5302 & 44.2 & 50.4 & 45.6 & 40.8
 &  240 & 32.2 & 50.6 & 35.6 & 23.1
 & 3191 & 28.5 & 34.9 & 29.6 & 25.0
 \tabularnewline
&  & StAR &  14881 & 0.6 & 12.9 & 0.6 & 0.1
 &  1126 & 6.2 & 14.4 & 5.9 & 2.3
 &  5055 & 8.4 & 10.4 & 8.4 & 7.1
 \tabularnewline

\cline{2-18}
  & \multirow{3}{*} {4} & TransE  & 7203 & 13.6 & 41.3 & 22.3 & 0.3 &  206 & 31.4 & 50.5 & 35.1 & 21.8 &  2766 & 25.1 & 32.7 & 26.3 & 21.2

\tabularnewline
&  & RotatE & 5278 & 44.0 & 50.2 & 45.3 & 40.7 &  244 & 32.3 & 50.5 & 35.6 & 23.2
 &  3183 & 28.6 & 34.9 & 29.6 & 25.2
 \tabularnewline
& & StAR &  14827 & 0.6 & 1.7 & 0.6 & 0.1 &  1467 & 4.4 & 9.9 & 3.6 & 1.7
 &  5158 & 7.8 & 9.6 & 8.0 & 6.5
 \tabularnewline
 \hline
 
 \multirow{18}{*} {ZeroE}  & \multirow{3}{*} {\makecell{5\\ w/ context}} & TransE  &  3598 & 7.6 & 12.1 & 7.9 & 5.1 &  787 & 11.8 & 19.9 & 12.4 & 7.7
  & 6170 & 0.8 & 1.5 & 0.7 & 0.3\tabularnewline
&  & RotatE &   4064 & 7.6 & 11.6 & 7.8 & 5.4 &  2412 & 2.6 & 4.9 & 2.5 & 1.1
 & 11872 & 0.2 & 0.4 & 0.2 & 0.1 \tabularnewline
&  & StAR &  487 & 36.3 & 58.0 & 41.0 & 25.5
 &  236 & 22.5 & 38.7 & 24.2 & 14.5
 &  1399 & 5.7 & 10.5 & 5.3 & 3.0
 \tabularnewline
\cline{2-18}

& \multirow{3}{*} {\makecell{5\\ w/o context}} & TransE  &  5966 & 2.3 & 4.5 & 2.3 & 1.0 & 876 & 10.9 & 17.9 & 11.2 & 7.1
  &  7314 & 0.5 & 1.0 & 0.4 & 0.1
\tabularnewline
&  & RotatE &  6492 & 1.9 & 3.8 & 1.7 & 0.9
 &  3037 & 1.0 & 1.7 & 0.8 & 0.4
 &  18656 & 0.0 & 0.0 & 0.0 & 0.0
 \tabularnewline
&  & StAR & \multicolumn{1}{c|}{-} & \multicolumn{1}{c|}{-} & \multicolumn{1}{c|}{-}& \multicolumn{1}{c|}{-}& \multicolumn{1}{c|}{-} &\multicolumn{1}{c|}{-} & \multicolumn{1}{c|}{-}& \multicolumn{1}{c|}{-} & \multicolumn{1}{c|}{-} & \multicolumn{1}{c|}{-} & \multicolumn{1}{c|}{-} & \multicolumn{1}{c|}{-}& \multicolumn{1}{c|}{-} & \multicolumn{1}{c|}{-} & \multicolumn{1}{c|}{-} \tabularnewline
\cline{2-18}

& \multirow{3}{*} {\makecell{6\\ w/ context}} & TransE  & 3605 &  7.5
 & 12.2 &  7.9 & 5.1 &  788 & 11.8 & 19.9 & 12.4 & 7.7
  &  6170 & 0.8 & 1.6 & 0.7 & 0.3
\tabularnewline
&  & RotatE &  4064 & 7.5 & 11.2 & 7.7 & 5.4
 &  2378 & 2.6 & 4.9 & 2.6 & 1.1
 &  16105 & 0.0 & 0.0 & 0.0 & 0.0
 \tabularnewline
&  & StAR &  481 & 34.7 & 56.6 & 39.3 & 23.9
 &  195 & 22.8 & 38.9 & 24.6 & 14.8
 &  1366 & 5.6 & 10.4 & 5.4 & 2.8
 \tabularnewline
\cline{2-18}

& \multirow{3}{*} {\makecell{6\\ w/o context}} & TransE  &  5981 & 2.3 & 4.5 &  2.2 & 1.0 &  876 & 10.9 & 17.9 & 11.2 & 7.1
  &  7314 & 0.5 & 1.0 & 0.4 & 0.1
\tabularnewline
&  & RotatE &  6510 & 1.9 & 3.7 & 1.8 & 0.8
 &  2997 & 1.0 & 1.7 & 0.8 & 0.4
 &  16056 & 0.0 & 0.0 & 0.0 & 0.0
 \tabularnewline
&  & StAR & \multicolumn{1}{c|}{-} & \multicolumn{1}{c|}{-} & \multicolumn{1}{c|}{-}& \multicolumn{1}{c|}{-}& \multicolumn{1}{c|}{-} &\multicolumn{1}{c|}{-} & \multicolumn{1}{c|}{-}& \multicolumn{1}{c|}{-} & \multicolumn{1}{c|}{-} & \multicolumn{1}{c|}{-} & \multicolumn{1}{c|}{-} & \multicolumn{1}{c|}{-}& \multicolumn{1}{c|}{-} & \multicolumn{1}{c|}{-} & \multicolumn{1}{c|}{-} \tabularnewline
\cline{2-18}

& \multirow{3}{*} {7} & TransE  & 3358 & 25.7 & 42.4 & 31.9 & 16.2 & 170 & 34.0 & 52.3 & 38.6 & 24.3  &  2475 & 17.4 & 25.4 & 18.5 & 13.2
\tabularnewline
&  & RotatE &  4427 & 3.5 & 43.6 & 37.5 & 30.6
 &  289 & 29.6 & 46.7 & 33.5 & 20.6
 &  14468 & 10.5 & 11.8 & 10.8 & 9.9
 \tabularnewline
&  & StAR & \multicolumn{1}{c|}{-} & \multicolumn{1}{c|}{-} & \multicolumn{1}{c|}{-}& \multicolumn{1}{c|}{-}& \multicolumn{1}{c|}{-} &\multicolumn{1}{c|}{-} & \multicolumn{1}{c|}{-}& \multicolumn{1}{c|}{-} & \multicolumn{1}{c|}{-} & \multicolumn{1}{c|}{-} & \multicolumn{1}{c|}{-} & \multicolumn{1}{c|}{-}& \multicolumn{1}{c|}{-} & \multicolumn{1}{c|}{-} & \multicolumn{1}{c|}{-} \tabularnewline

\cline{2-18}
& \multirow{3}{*} {8} & TransE  & 3887 &  25.3 & 41.6 & 31.0 & 16.2 &  184 & 33.8 & 52.1 & 38.5 & 24.1
 & 3673 & 15.7 & 23.3 & 16.5 & 11.9
\tabularnewline
&  & RotatE &  4475 & 35.2 & 43.7 & 37.3 & 30.4
 &  248 & 29.8 & 47.3 & 33.4 & 20.7
 &  15469 & 6.7 & 8.9 & 7.5 & 5.5
\tabularnewline
&  & StAR & \multicolumn{1}{c|}{-} & \multicolumn{1}{c|}{-} & \multicolumn{1}{c|}{-}& \multicolumn{1}{c|}{-}& \multicolumn{1}{c|}{-} &\multicolumn{1}{c|}{-} & \multicolumn{1}{c|}{-}& \multicolumn{1}{c|}{-} & \multicolumn{1}{c|}{-} & \multicolumn{1}{c|}{-} & \multicolumn{1}{c|}{-} & \multicolumn{1}{c|}{-}& \multicolumn{1}{c|}{-} & \multicolumn{1}{c|}{-} & \multicolumn{1}{c|}{-} \tabularnewline
 \hline
 
 \multirow{6}{*} {ZeroR}  & \multirow{3}{*} {9} & TransE  & 19585 &   0.3 & 0.4 & 0.2 &  0.1 &   6922 & 2.7 & 4.3 & 2.8 & 1.7

  & 7799 &0.5  & 0.8 & 0.2& 0.1\tabularnewline
&  & RotatE &  21877 & 0.2 & 0.4 & 0.2 & 0.1
 &  6645 & 1.6 & 2.3 & 1.6 & 1.2
 &  14468 & 10.5 & 11.8 & 10.7 & 9.9
 \tabularnewline
&  & StAR &  2810 & 9.7 & 24.6 & 11.1 & 2.4
 &  1836 & 11.8 & 21.2 & 12.1 & 7.0
 & 2787 & 2.6 & 4.3 & 2.3 & 1.4
 \tabularnewline
\cline{2-18}

& \multirow{3}{*} {10} & TransE  & 22324 & 0.1 & 0.1 & 0.1 & 0.1 &    7037 & 2.7 & 4.1 & 2.9 & 1.8 &
  10018 & 0.4 & 0.8 & 0.2 & 0.0
\tabularnewline
&  & RotatE &  18464 & 0.4 & 0.6 & 0.4 & 0.1
 &  3579 & 2.7 & 5.4 & 2.5 & 1.2
 &  18562 & 0.1 & 0.1 & 0.0 & 0.0
 \tabularnewline
&  & StAR &  15186 & 0.9 & 2.3 & 0.9 & 0.3
 &  3903 & 1.5 & 4.1 & 1.0 & 0.5 &  8514 & 2.0 & 3.1 & 1.9 & 1.4
\tabularnewline

 \hline
 
 \multirow{9}{*} {ZeroB}  & \multirow{3}{*} {\makecell{11\\ w/ context}} & TransE  & 4830 & 4.0 & 9.0 & 3.2 & 2.1 & 3581 & 3.8 & 6.9 & 4.2 & 1.9
& 15032 & 2.2 & 3.8 & 2.0& 0.1 \tabularnewline
&  & RotatE & 7116 & 4.3 & 8.9 & 4.1 & 2.6
 &  3855 & 1.5 & 3.8 & 1.1 & 0.4
 &  10405 & 1.8 & 1.9 & 1.6 & 1.6 \tabularnewline
&  & StAR & 5317 & 14.1 & 29.6 & 15.4 & 7.7
 &   1561 & 10.4 & 19.2 & 10.5 & 5.8

 &  2102 & 2.3 & 4.1 & 2.0 & 1.0
 \tabularnewline
\cline{2-18}

& \multirow{3}{*} {\makecell{11\\ w/o context}}& TransE  & 8664 & 1.7 & 3.0 & 1.6 & 0.6 &  5425 & 0.9 & 1.4 & 0.9 & 0.5

  & 22389 & 0.0 & 0.0 & 0.0 & 0.0
\tabularnewline
&  & RotatE &  11949 & 0.0 & 0.0 & 0.0 & 0.0

&  6326 & 0.1 & 0.1 & 0.0 & 0.0
 &  15049 & 0.1 & 0.1 & 0.0 & 0.0
 \tabularnewline
&  & StAR & \multicolumn{1}{c|}{-} & \multicolumn{1}{c|}{-} & \multicolumn{1}{c|}{-}& \multicolumn{1}{c|}{-}& \multicolumn{1}{c|}{-} &\multicolumn{1}{c|}{-} & \multicolumn{1}{c|}{-}& \multicolumn{1}{c|}{-} & \multicolumn{1}{c|}{-} & \multicolumn{1}{c|}{-} & \multicolumn{1}{c|}{-} & \multicolumn{1}{c|}{-}& \multicolumn{1}{c|}{-} & \multicolumn{1}{c|}{-} & \multicolumn{1}{c|}{-} \tabularnewline
\cline{2-18}

& \multirow{3}{*} {12} & TransE  & 15125 & 0.0 & 0.0 & 0.0 & 0.0 & 3167 & 6.4 & 10.7 & 7.0 & 3.9  & 14534 & 2.6 & 2.7 & 2.6 & 2.5
\tabularnewline
&  & RotatE & 14867 & 0.0 & 0.0 & 0.0 & 0.0 &  5424 & 0.3 & 0.4 & 0.3 & 0.1
 &  15089 & 0.1 & 0.1 & 0.1 & 0.1
 \tabularnewline
&  & StAR & \multicolumn{1}{c|}{-} & \multicolumn{1}{c|}{-} & \multicolumn{1}{c|}{-}& \multicolumn{1}{c|}{-}& \multicolumn{1}{c|}{-} &\multicolumn{1}{c|}{-} & \multicolumn{1}{c|}{-}& \multicolumn{1}{c|}{-} & \multicolumn{1}{c|}{-} & \multicolumn{1}{c|}{-} & \multicolumn{1}{c|}{-} & \multicolumn{1}{c|}{-}& \multicolumn{1}{c|}{-} & \multicolumn{1}{c|}{-} & \multicolumn{1}{c|}{-} \tabularnewline

 \hline
 
\end{tabular}

\label{tab:full_results}
\end{table*}

\section{Experiments}
We next evaluate three canonical KBC systems on our proposed ZeroKBC benchmark. As these systems are not originally designed to handle zero-shot scenarios, we adapt them accordingly to make predictions under zero-shot settings.

\subsection{Systems}

\noindent \textbf{TransE}~\cite{bordes2013translating} embeds entities and relations into low-dimension vectors and interprets relations $r$ as translations operating on the head $h$ and tail $t$ entities. The model tries to minimize the distance $d=||h+r-t||$ with $h,r,t \in \mathbb{R}^{k}$. In its vanilla settings, the initial embeddings of entities and relations are random; in our adaptation, we use BERT~\cite{devlin-etal-2019-bert} to encode the textual descriptions of entities and relations for initialization when the descriptions are available\footnotemark[3]. In this case, unseen entities or relations will use the BERT embeddings for prediction. When the context is available, we use the translation property to deduce the embeds for unseen entities\footnotemark[4], as $h_{\text{unseen}} = t_{\text{seen}} - r$ or  $t_{\text{unseen}} = h_{\text{seen}} + r$. 


\footnotetext[3]{We regard each description as a sentence and then use the embedding of [CLS] as the entity/relation representation.}

\footnotetext[4]{For an unseen entity that has multiple context connections, we average them.}

\noindent \textbf{RotatE}~\cite{sun2018rotate} embeds entities and relations into complex space vectors and interprets relations $r$ as rotations operating in the space. Similar to TransE, we use BERT to initialize the real part of the entities and relations. For
unseen entities, we use $h \circ r = t$ and $h,r,t \in \mathbb{C}^{k}$ for deductions.  


\noindent \textbf{StAR}~\cite{wang2021structure} adopts BERT as its textual encoder and uses both the translation property of a triple and their textual semantics that can be more generalizable for unseen entities and relations.\footnotemark[5]

\footnotetext[5]{Because the model regards the head entity and relation as a whole, we cannot deduce to unseen entities using context.}

\subsection{Results and Analysis}
We report the main experimental results in Table \ref{tab:main_results} when all possible information is available for each scenario (corresponding to the first row of each colored scenario in Figure \ref{fig:task}). Specifically, we assume that the model always has the textual description of newly emerged entities or relations for ZeroE or ZeroR, respectively. 


Compared with the standard KBC setting, the performance of canonical KBC systems (i.e., TransE, RotatE, and StAR) drops dramatically under zero-shot entity or relation setting, demonstrating the difficulty of our proposed dataset. ZeroB is the most challenging setting since both entities and relations in test data are unseen during training. Additionally, StAR can encode the semantics of text descriptions more efficiently, which significantly outperforms TransE and RotatE across knowledge graph genres and zero-shot settings. It indicates \textbf{the importance of considering textual semantics in zero-shot KBC} and points to the future direction about how to design an efficient approach that can take descriptive information into consideration to solve Zero-KBC problems. In our experiments, we also observe that \textbf{graph properties of knowledge bases can affect models' generalization capabilities}. For example, compared with WN18RR and FB15K237, ATOMIC has a much sparser graph structure (only 1\% of the entities act as both the head and tail, while the percentage for WN18RR and FB15K237 are 80\% and 88\%, respectively).
As a result, the models that only leverage the graph structures suffer from the largest performance drop from the standard setting to zero-shot ones, while the model (i.e., StAR) utilize it could generalize better to unseen entities and relations.
We also include complete experimental results on all ZeroKBC scenarios in Table \ref{tab:full_results}.


\section{Conclusion}

This work systematically examines different zero-shot KBC scenarios and develops a comprehensive benchmark named ZeroKBC, which covers these scenarios with three types of knowledge sources. Experimental results show that canonical and state-of-the-art KBC systems suffer great performance degradation on this challenging yet practical benchmark. We further present several important observations and promising future directions based on our system output analysis.                  

\bibliographystyle{IEEEtran}
\bibliography{custom.bib}{}

\end{document}